\documentclass[11pt]{article}

\usepackage[final]{acl}

\usepackage{times}
\usepackage{latexsym}

\usepackage[T1]{fontenc}

\usepackage[utf8]{inputenc}

\usepackage{microtype}

\usepackage{inconsolata}

\usepackage{graphicx}

\usepackage{amsmath}
\usepackage{booktabs}
\usepackage{multirow}

\usepackage{soul}
\renewcommand{\st}[1]{}
\definecolor{darkgreen}{rgb}{0.0, 0.8, 0.0}
\definecolor{darkyellow}{rgb}{0.85, 0.65, 0.13}
\definecolor{darkred}{rgb}{0.8, 0.0, 0.0}
\ifodd 0
\newcommand{\ws}[1]{{\color{darkred} #1}}
\newcommand{\XR}[1]{{\color{darkgreen} #1}}
\else
\newcommand{\XR}[1]{#1} 
\newcommand{\ws}[1]{#1}   
\fi 

\ifodd 0
\newcommand{\FV}[1]{{\color{darkyellow}#1}}
\else
\newcommand{\FV}[1]{#1}
\fi

\title{Visual Attention Faithfulness in Vision-Language Models is Heterogeneous}

\author{
  Xurui Song\textsuperscript{1,2}\thanks{Work done during internship at SAP.}, \quad
  Weishi Wang\textsuperscript{1}\thanks{Corresponding author.}, \quad
  Zhongqi Yue\textsuperscript{3}, \quad
  Kuluhan Binici\textsuperscript{1} \\
  {\bfseries Tao Bai\textsuperscript{1}, \quad
  Hongxin Shao\textsuperscript{1}, \quad
  Daniel Dahlmeier\textsuperscript{1}, \quad
  Jun Luo\textsuperscript{2}} \\[0.6em]
  \textsuperscript{1}SAP \\
  \textsuperscript{2}College of Computing and Data Science, Nanyang Technological University, Singapore \\
  \textsuperscript{3}Microsoft
}

\begin{document}
\maketitle

\begin{abstract}
Whether attention weights faithfully reflect model reasoning has been actively debated in NLP, yet this question remains largely unexplored for the visual modality in Vision-Language Models (VLMs). We address this gap through causal perturbation analysis on current VLMs, evaluating both the comprehensiveness and sufficiency gap of attention-ranked visual tokens. Our analysis reveals that visual attention faithfulness is heterogeneous, manifesting in three distinct processing modes: \textit{Faithful-Sufficient}, where top-$k$ attention tokens are both necessary and sufficient for prediction; \textit{Faithful-Distributed}, where they are necessary but broader visual context remains required; and \textit{Non-Focal}, where no localized attention region is individually necessary while visual information remains an essential trigger for prediction. Furthermore, human-annotated ground-truth regions satisfy comprehensiveness in only $\sim 60$\% of cases compared with model attention rankings, revealing systematic divergence between model visual reliance and human intuition. We demonstrate these patterns across both general VQA on VQAv2 and document 
\FV{tasks} on VRDU and ChartQA, showing that visual attention faithfulness varies systematically with processing demands and model architectures rather than being uniformly faithful or unfaithful.
\end{abstract}

\section{Introduction}

\begin{figure*}[t]
\centering
\includegraphics[width=\textwidth]{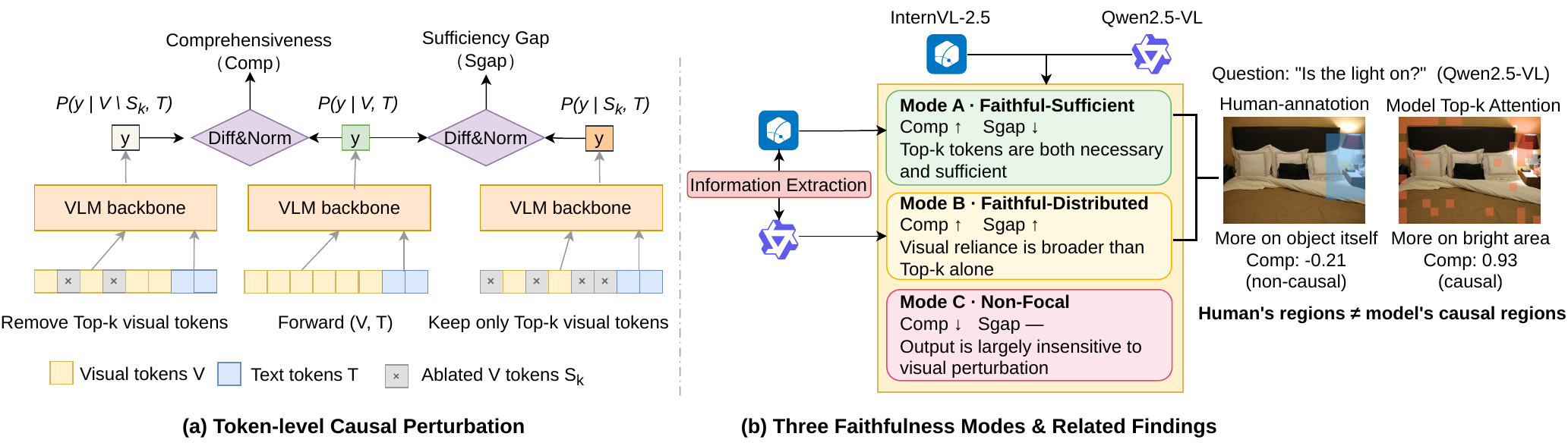}
\caption{Overview of the proposed evaluation framework and main findings. (Left) Visual attention faithfulness is evaluated using comprehensiveness and the sufficiency gap under token-level perturbations, by respectively removing or retaining top-$k$ attention-ranked visual tokens. (Right) Three faithfulness modes of visual attention of the VLM backbone. Different VLM architectures may handle the same task in different modes, for example, the IE task. Furthermore, what humans perceive as causal regions is not always the focus of the model's attention.}
\label{fig:overview}
\vspace{-0.1em}
\end{figure*}
Despite the widespread use of attention weights as a window into model reasoning \cite{clark2019bert,chefer2021generic}, whether attention provides a faithful explanation of how models arrive at their predictions remains an open question. In text-based Natural Language Processing (NLP), this question has sparked sustained debate: evidence suggests that attention distributions can be permuted without changing predictions \cite{jain2019attention}, while counter-evidence shows that models with constrained attention behave meaningfully differently \cite{wiegreffe2019attention}. A key insight from this line of work is that faithfulness should not be treated as 
\ws{a binary property of attention but rather a graded, intervention-specific property that has to be measured by sample} \cite{jacovi2020faithfully,deyoung2020eraser}. Yet 
this faithfulness debate has been developed in text-only settings, leaving open whether its conclusions transfer to architectures that jointly process visual and textual information.

Vision-Language Models (VLMs) \cite{liu2024llava,chen2024internvl25,bai2025qwen25vl,openai2024gpt4o} \FV{consisting of a vision transformer encoder~\cite{ViT} and a language backbone~\cite{GPT-1}} represent precisely such an architecture. By encoding image patches as visual tokens processed alongside text within a unified transformer \cite{vaswani2017attention}, VLMs create a setting where attention operates over inputs with 2D spatial structure and high inter-token redundancy among neighboring patches. However, existing work on VLM attention has proceeded without first establishing whether attention faithfully reflects the model's reliance on visual information. Recent studies either assume that low-attention visual tokens are dispensable for inference acceleration \cite{chen2024fastv,shang2024prumerge}, or analyze attention patterns to characterize internal mechanisms \cite{golovanevsky2024notice,jiang2024devils}, yet the fundamental question of whether visual attention rankings correspond to actual model dependencies on specific visual regions remains unexplored.

\XR{We address this question through token-level causal perturbation analysis on representative open-source VLMs, 
\FV{Qwen2.5-VL} and InternVL2.5 \cite{bai2025qwen25vl,chen2024internvl25}. As illustrated in Figure~\ref{fig:overview}, we ablate visual token embeddings at the first decoder layer, either removing or retaining tokens according to their attention ranking, and measure the resulting change in prediction probability. We evaluate both comprehensiveness and sufficiency gap of attention-ranked visual tokens on VQAv2 \cite{goyal2017vqav2}. This evaluation captures whether attention identifies visual tokens that are necessary for the prediction, or whether they alone are sufficient to recover it.
}

Our analysis reveals that visual attention faithfulness is not uniform but heterogeneous in VLMs, manifesting in three distinct processing modes. In the \textit{Faithful-Sufficient} mode 
, attention top-$k$ tokens are both necessary and sufficient, indicating that attention captures the visual information the model relies on for these predictions. In the \textit{Faithful-Distributed} mode 
, top-$k$ tokens are necessary but not sufficient, as the model draws on broader visual context beyond its focal attention region. In the \textit{Non-Focal} 
, no concentrated attention region is necessary for the output, yet visual information overall remains an essential trigger for prediction, indicating diffusely distributed visual dependencies that attention rankings cannot localize. 


Next, we compare model visual attention against human-annotated ground-truth regions indicating which areas humans consider task-relevant. This comparison reveals systematic divergence: ground-truth regions satisfy comprehensiveness in only $\sim$60\% of cases 
, indicating that models develop visual processing strategies that 
and diverge from the regions humans deem important. 

\XR{We further validate these patterns on document information extraction (IE) using VRDU~\cite{wang2023vrdu} dataset \FV{and document VQA using ChartQA~\cite{chartqa}, practical tasks where VLMs must locate, extract or understand field values} from visually structured documents. \FV{On these tasks}, both Qwen2.5-VL and InternVL2.5 reliably attend to answer-relevant regions, yet exhibit different processing modes. Qwen2.5-VL behaves as an extreme instance of the \textit{Faithful-Distributed} mode, where evidence regions are highly necessary but insufficient without broader layout context. In contrast, InternVL2.5 behaves closer to the \textit{Faithful-Sufficient} mode, where attended regions are both necessary and largely sufficient for prediction. These findings show that the three processing modes extend beyond general VQA to structurally different visual tasks, while also suggesting that different VLM architectures may adopt different visual processing strategies for the same task.} Our contributions are as follows:
\begin{itemize}
\item We present the first systematic evaluation of visual attention faithfulness in VLMs, measuring both comprehensiveness and sufficiency via causal perturbation.
\item \XR{We identify three visual attention processing modes that characterize how VLMs rely on visual information, showing that faithfulness is heterogeneous rather than uniform.}
\item We reveal systematic divergence between model visual attention and human-annotated ground-truth regions, showing that human annotations are an incomplete proxy for model visual dependencies.
\item \XR{We further show that these modes generalize beyond VQA to document 
\FV{understanding tasks} and vary across VLM architectures.}
\end{itemize}

\section{Related Work}

\paragraph{Attention as Explanation.}
Prior work on attention interpretability in NLP has produced conflicting conclusions. While \citet{jain2019attention} and \citet{serrano2019attention} argue that attention weights show weak correspondence with feature importance and can be manipulated without affecting predictions, \citet{wiegreffe2019attention} contend that these findings do not invalidate attention as an explanation, since constraining attention during training leads to measurably different model behavior. \citet{bastings2020elephant} further questions attention's role by demonstrating that gradient-based saliency methods are better at identifying causal tokens. Meanwhile, attention patterns have been productively used to probe model internals, revealing that specific heads encode syntactic and semantic relations \cite{clark2019bert,vig2019multiscale,michel2019sixteen}. These works establish that attention occupies an ambiguous position between faithful explanation and mere computational byproduct, with conclusions drawn exclusively from text-only architectures.

\paragraph{Faithfulness Evaluation.}
To move beyond qualitative debates, several frameworks have been proposed for quantitatively measuring explanation faithfulness. \citet{jacovi2020faithfully} formalize faithfulness as a graded property requiring empirical verification rather than assumption. \citet{deyoung2020eraser} operationalize this through comprehensiveness and sufficiency metrics that measure whether highlighted tokens are necessary or sufficient for the prediction, and \citet{madsen2022faithfulness} propose recursive masking and retraining for faithfulness assessments. For attention aggregation specifically, \citet{abnar2020quantifying} propose attention rollout and attention flow to aggregate raw attention across layers, addressing that single-layer attention may not reflect overall information routing. These evaluation paradigms provide the methodological foundation for systematically probing attention faithfulness, though they have been developed and validated exclusively on text inputs.

\paragraph{Interpretability in VLMs.}
Interpretability research on Vision-Language Models has pursued several directions. Visual explanation methods such as Grad-CAM \cite{selvaraju2017gradcam} and attention-based approaches \cite{chefer2021generic,aflalo2022vlinterpret} generate spatial heatmaps highlighting image regions relevant to predictions, but do not verify whether these highlighted regions causally drive model outputs. On the mechanistic side, \citet{golovanevsky2024notice} identifies universal cross-attention heads responsible for specific functions like object detection, while \citet{jiang2024devils} link attention patterns in middle layers to hallucination phenomena. Recent work has also identified attention sink phenomena in both language models \cite{xiao2024streamingllm} and vision transformers \cite{darcet2024registers}, where certain tokens accumulate disproportionate attention regardless of semantic content. A separate line of work leverages attention sparsity for efficiency, assuming that visual tokens with low attention scores can be safely pruned or merged \cite{chen2024fastv,shang2024prumerge,zhang2025sparsevlm,bolya2023tome}. However, no prior work has systematically measured whether attention-ranked visual tokens are causally necessary or sufficient for VLM predictions. Our work fills this gap by applying comprehensiveness and sufficiency evaluation to VLM visual attention, revealing heterogeneous faithfulness patterns that these prior approaches do not account for.

\section{Method}

We formalize the evaluation of visual attention faithfulness in VLMs as a causal measurement problem. Given a VLM that processes both visual and textual tokens, we ask: do the visual tokens that receive the highest attention weights causally correspond to the tokens the model relies on for its predictions? We operationalize this question through three components: attention-based token ranking (\S\ref{sec:ranking}), causal perturbation (\S\ref{sec:perturbation}), and faithfulness metrics (\S\ref{sec:metrics}).

\subsection{Problem Formulation}

Consider a VLM that takes an image $I$ and a text query $Q$ as input. The image is encoded into a sequence of visual tokens $\mathbf{V} = \{v_1, v_2, \ldots, v_N\}$, and the query into text tokens $\mathbf{T} = \{t_1, t_2, \ldots, t_M\}$. The model autoregressively generates an answer 
\FV{$\mathbf{y} = (y_1, \ldots, y_{|\mathbf{y}|})$}, with $P(\mathbf{y} \mid \mathbf{V}, \mathbf{T})$ denoting the mean probability assigned to the generated answer tokens. Our goal is to determine whether the attention-derived importance ranking over visual tokens faithfully reflects their causal contribution to $P(\mathbf{y} \mid \mathbf{V}, \mathbf{T})$.

\subsection{Attention-based Token Ranking}
\label{sec:ranking}

To obtain a single importance score for each visual token, we aggregate attention weights across all layers and heads. In a transformer with $L$ layers and $H$ attention heads per layer, let $\alpha_{l,h}(i, j)$ denote the attention weight from token $j$ to token $i$ at layer $l$, head $h$. 
We compute the aggregated attention score for each visual token $v_i$ as:
\begin{equation}
a_i = \sum_{l=1}^{L} \frac{1}{H} \sum_{h=1}^{H} \frac{1}{|\mathbf{y}|} \sum_{s=1}^{|\mathbf{y}|} \alpha_{l,h}(i, y_s).
\end{equation}
We average over heads as an unbiased aggregation since the dominant routing heads may vary across layers and inputs \cite{michel2019sixteen}, and \FV{summing across layers captures the native attention distribution across the network without additional propagation assumptions, making it suitable for assessing visual attention faithfulness.} Sorting visual tokens by $a_i$ in descending order and selecting the top-$k$ yields $\mathcal{S}_k = \{v_{(1)}, v_{(2)}, \ldots, v_{(k)}\}$, where $(i)$ denotes the index with the $i$-th largest score.

\subsection{Causal Perturbation}
\label{sec:perturbation}

To establish a causal rather than merely correlational link between attention rankings and model predictions, we adopt an interventionist approach: selectively ablating visual tokens and observing the effect on prediction probability. Specifically, for a given token set $\mathcal{S}$, we zero-ablate their hidden states $\mathbf{h}_i^{(0)}$ at the entry of the first decoder layer:
\begin{equation}
\mathbf{h}_i^{(0)} \leftarrow \mathbf{0}, \quad \forall\, v_i \in \mathcal{S}.
\end{equation}
This zero-ablation introduces no additional information into the representation space, removing the contribution of the ablated tokens from all subsequent computations.

\subsection{Faithfulness Metrics}
\label{sec:metrics}

We evaluate attention faithfulness along two complementary dimensions adapted from \citet{deyoung2020eraser}: comprehensiveness measures whether attention-highlighted tokens are \emph{necessary}, while sufficiency gap measures whether they are \emph{sufficient}. Since different samples have varying baseline prediction probabilities $P(\mathbf{y} \mid \mathbf{V}, \mathbf{T})$, we normalize both metrics by this value to ensure comparability across samples.

\paragraph{Comprehensiveness.} Removing the top-$k$ attention-ranked tokens should degrade prediction if they are causally important:
\begin{equation}
\text{Comp}(\mathcal{S}_k) \!=\! \frac{P(\mathbf{y} \mid \mathbf{V}, \mathbf{T}) \!-\! P(\mathbf{y} \mid \mathbf{V} \!\setminus\! \mathcal{S}_k, \mathbf{T})}{P(\mathbf{y} \mid \mathbf{V}, \mathbf{T})}.
\end{equation}
High comprehensiveness indicates that the top-$k$ tokens are necessary for the prediction.

\paragraph{Sufficiency gap.} Retaining only the top-$k$ tokens should preserve prediction if they capture all necessary visual information. We define the sufficiency gap (Sgap) as the normalized probability drop after this restriction:
\begin{equation}
\text{Sgap}(\mathcal{S}_k) = \frac{P(\mathbf{y} \mid \mathbf{V}, \mathbf{T}) - P(\mathbf{y} \mid \mathcal{S}_k, \mathbf{T})}{P(\mathbf{y} \mid \mathbf{V}, \mathbf{T})}.
\end{equation}
A low (or negative) sufficiency gap indicates that the top-$k$ tokens alone are sufficient for the prediction. Together, these two metrics locate each sample in a comprehensiveness--Sgap space that characterizes the degree and nature of attention faithfulness.

\section{Experiments and Analysis}

\subsection{Experiment Setup}
\label{sec:setup}

\paragraph{Model and Inference.}
\FV{We conduct our main experiments on Qwen2.5-VL-7B, evaluate the generalizability of our findings on InternVL2.5-8B, and further perform scaling experiments with Qwen2.5-VL-3B and InternVL2.5-4B.}
We use greedy decoding (temperature = 0) throughout to ensure deterministic outputs, so that any changes in $P(\mathbf{y} \mid \mathbf{V}, \mathbf{T})$ can be directly attributed to token ablations rather than sampling variance.

\paragraph{Datasets.}
We evaluate on two benchmarks that exercise different aspects of visual understanding.
For general visual question answering (VQA), we sample 90 image-question pairs from VQAv2 \cite{goyal2017vqav2}, balanced across three question categories (30 yes/no, 30 number, 30 other) to capture a range of reasoning demands from binary judgments to open-ended descriptions. Each sample is accompanied by a manually verified ground-truth (GT) region indicating the image area that humans consider task-relevant (details in Appendix~\ref{app:setup}).
For document 
IE task, we randomly choose 
\FV{60} samples \FV{answered correctly} from the VRDU 
\cite{wang2023vrdu} spanning two document types (advertisement and registration forms), \FV{and we select 100 samples from the ChartQA~\cite{chartqa} test set for the document VQA task.}

\paragraph{Comparison Strategies.}
To isolate the causal contribution of attention-ranked tokens, for VQA we compare four perturbation strategies applied to the same number of tokens $k$:
\begin{itemize}
\item \textbf{Overall Top-K}: the $k$ visual tokens receiving the highest aggregated attention scores $a_i$ (\S\ref{sec:ranking}), representing the model's own importance ranking.
\item \textbf{GT Object}: the $k$ tokens corresponding to the human-annotated GT region, representing human judgment of task-relevant areas.
\item \textbf{Non-GT Top-K}: the $k$ highest-attention tokens that fall \emph{outside} the GT region, capturing attention to non-obvious regions.
\item \textbf{Random}: $k$ tokens randomly sampled at fixed seed, serving as an uninformed baseline.
\end{itemize}
The token count $k$ is set equal to the number of visual tokens covered by the GT region for each sample, ensuring that all four strategies ablate the same proportion of visual information. This alignment makes comprehensiveness and sufficiency gap scores directly comparable across strategies: any observed differences reflect the importance of the \emph{selected} tokens rather than an artifact of varying ablation size. For IE, instead of GT-aligned comparison, we ablate the top 1\%, 3\%, 5\%, and 10\% of visual tokens by attention score to trace the progression of causal effect as coverage grows from the most focal to broader regions. This is because model attention on these documents is already highly localized to the answer regions (\S\ref{sec:ie}), making a GT-aligned comparison largely redundant, and document layouts contain redundant copies of the same information across regions, which a single GT mask cannot capture. \FV{For the CharQA task, we adopt a similar proportion perturbation as IE for comparison.}

\paragraph{Evaluation Protocol.}
For each strategy and each sample, we compute comprehensiveness by zero-ablating the selected tokens and measuring the normalized probability drop, and the sufficiency gap by retaining only the selected tokens while ablating all others (\S\ref{sec:metrics}). Further implementation details and visual strategies of the tested models
are described in Appendix~\ref{app:setup}.

\subsection{Three Processing Modes}
\label{sec:three-modes}

Excluding 5 samples that need whole image understanding
(Appendix~\ref{app:excluded-samples}), Table~\ref{tab:strategy-aggregate} reports the results of the four perturbations on the remaining valid samples. Among the four strategies, overall top-$k$ achieves the highest mean comprehensiveness (0.813) and the lowest mean sufficiency gap (0.469), suggesting that attention-ranked tokens are both highly necessary and sufficient for the model’s predictions, and that visual attention partially reflects the model’s reliance on visual input.

\begin{table}[t]
\centering
\small
\begin{tabular}{lcc}
\toprule
Strategy & Mean Comp & Mean Sgap \\
\midrule
Overall Top-$k$ & \textbf{0.813} & \textbf{0.469} \\
GT Object       & 0.505 & 0.846 \\
Non-GT Top-$k$  & 0.659 & 0.653 \\
Random          & 0.502 & 0.599 \\
\bottomrule
\end{tabular}
\caption{Mean normalized comprehensiveness (Comp) and sufficiency gap (Sgap) of the four perturbation strategies on the valid VQA samples. Each strategy ablates the same number of visual tokens $k$, equal to the GT region size for the corresponding sample.}
\label{tab:strategy-aggregate}
\end{table}

\begin{figure*}[t]
\centering
\includegraphics[width=\textwidth]{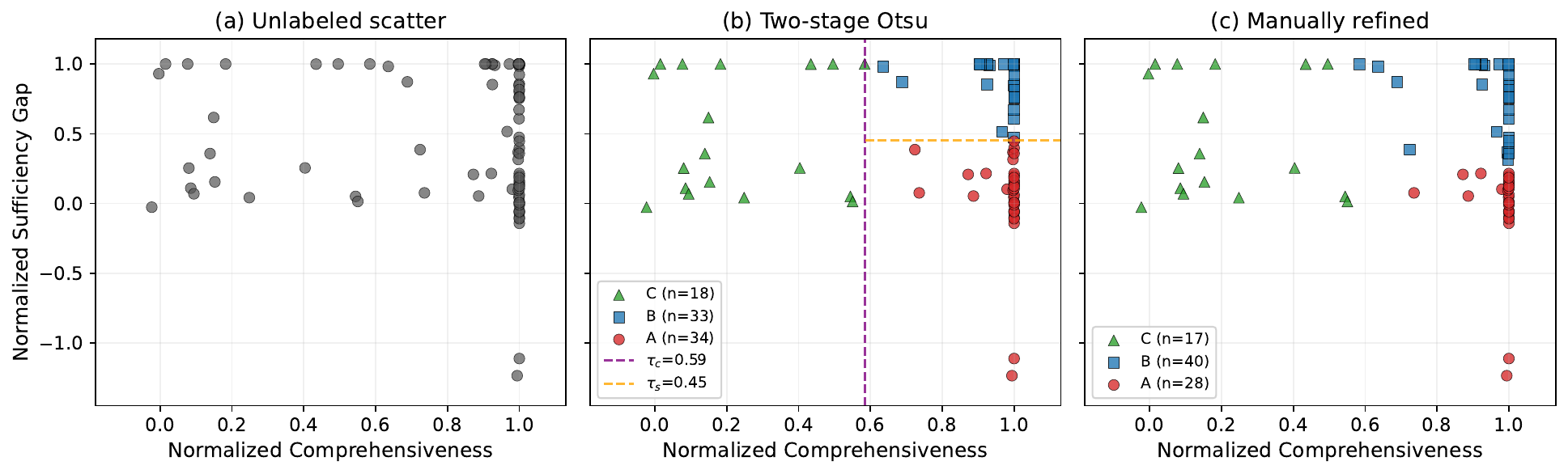}
\caption{Per-sample distribution of normalized comprehensiveness and sufficiency gap on the valid VQA samples. (a) Unlabeled scatter. (b) Two-stage Otsu thresholding yields a three-way partition. (c) Manually refined partition used in the analysis.}
\label{fig:processing-modes-scatter}
\end{figure*}

We then plot valid samples' overall top-$k$ attention tokens in the (Comp,Sgap) space in Figure~\ref{fig:processing-modes-scatter}.
The distribution exhibits a bimodal structure along the comprehensiveness axis, with the high-Comp region further separating into two bands of sufficiency gaps. Since comprehensiveness identifies whether a localized causal contributor exists, whereas sufficiency is only meaningful within such focal samples, we apply Otsu’s method~\cite{otsu1979threshold} sequentially: first on Comp over all samples, and then on Sgap within the high-Comp subset. This produces the three-way partition shown in Figure~\ref{fig:processing-modes-scatter}b. We further inspect samples near the decision boundaries and manually refine a small number of assignments based on their attention maps and prediction behavior. The resulting partition (Figure~\ref{fig:processing-modes-scatter}c) defines the three groups used throughout the remainder of the analysis: \emph{Faithful-Sufficient} (A), \emph{Faithful-Distributed} (B), and \emph{Non-Focal} (C). \FV{Across three modes, the model maintains consistently high answer accuracy (Table~\ref{tab:mode-summary}), suggesting the observed partition reflects genuine differences in the model's reliance on visual attention rather than being affected by incorrectness.}

\paragraph{Mode A: Faithful-Sufficient.}

\begin{figure*}[t]
\centering
\includegraphics[width=0.985\textwidth]{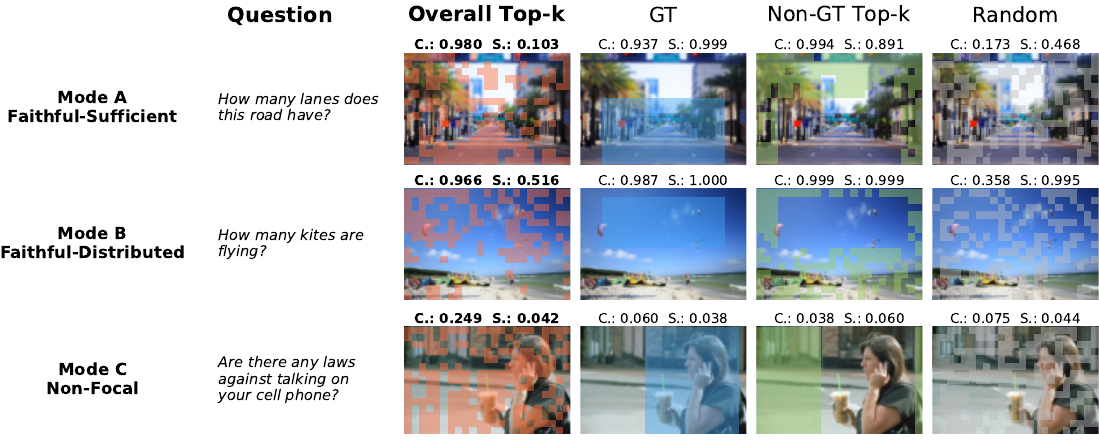}
\caption{Representative samples for the three processing modes. Each row visualizes the four perturbation strategies on a representative case and reports the per-strategy normalized comprehensiveness (\textbf{C.}) and sufficiency gap (\textbf{S.}), illustrating why the top-$k$ tokens are sufficient in Mode A, only necessary in Mode B, and almost inert in Mode C.}
\label{fig:representative-samples}
\end{figure*}

Mode A accounts for 32.9\% of samples and is characterized by near-maximal Comp ($0.978$) together with a slightly negative Sgap ($-0.026$) (Table~\ref{tab:mode-summary}), indicating that the top-$k$ attention tokens are both causally necessary and individually sufficient for the prediction, while the remaining tokens contribute little or act as mild distractors. This mode is typically associated with questions answerable from localized object-centric evidence without requiring broader spatial reasoning. A representative example asks for the number of lanes on a road, where the top-$k$ tokens align with the lane markings; ablating them collapses the prediction, whereas retaining only them preserves it (Figure~\ref{fig:representative-samples}, top row). In this regime, high-attention regions closely correspond to the evidence underlying the model’s prediction.

\paragraph{Mode B: Faithful-Distributed.}

Mode B accounts for 47.1\% of samples and is characterized by high Comp ($0.952$) together with a large Sgap ($0.818$) (Table~\ref{tab:mode-summary}), indicating that the top-$k$ attention tokens are causally important but insufficient on their own to recover the prediction. Questions in this mode typically require integrating localized evidence with broader spatial or contextual information. For example, in the question \textit{How many kites are flying?}, the top-$k$ tokens concentrate on the kite cluster and remain highly necessary for the prediction, yet retaining only them leaves a substantial sufficiency gap, suggesting that accurate counting additionally depends on surrounding scene context (Figure~\ref{fig:representative-samples}, middle row). In this regime, attention faithfully identifies relevant evidence, but does not fully capture the distributed visual information underlying the model’s decision.

\paragraph{Mode C: Non-Focal.}

Mode C accounts for 20.0\% of samples and is characterized by low Comp ($0.213$) together with a moderate Sgap ($0.460$), indicating that the top-$k$ attention tokens contribute little to the prediction despite the model still relying on visual input. Full-image ablation yields a Comp of $0.973$ \FV{while removing all task-irrelevant text tokens such as format instructions only yields a Comp of $0.306$}, confirming that the image remains necessary even though no localized attention region is decisive. Questions in this mode exhibit limited reliance on specific visual evidence, with the image primarily serving as a contextual co-activation trigger. For example, in a question about laws against cellphone use, the image establishes the cellphone context that activates the model’s relevant internal knowledge (Figure~\ref{fig:representative-samples}, bottom row). In this mode, visual attention is not faithful enough, but is indispensable context triggers.

\begin{table}[t]
\centering
\small
\setlength{\tabcolsep}{5pt}
\begin{tabular}{lcccc}
\toprule
Mode & \% & Comp & Sgap & Acc\\
\midrule
A: Faithful-Sufficient   & 32.9 & 0.978 & $-0.026$ & 92.9 \\
B: Faithful-Distributed  & 47.1 & 0.952 & 0.818 & 100.0\\
C: Non-Focal             & 20.0 & 0.213 & 0.460 & 94.1\\
\bottomrule
\end{tabular}
\caption{\FV{Per-mode share, mean overall top-$k$ scores and model accuracy (\%)\protect\footnotemark{} on the valid VQA samples.}}
\label{tab:mode-summary}
\end{table}
\footnotetext{Accuracy per mode is assessed by VQAv2 ground truth.}

\begin{table}[t]
\centering
\small
\setlength{\tabcolsep}{6pt}
\begin{tabular}{lccc}
\toprule
Question Type & A & B & C \\
\midrule
Yes/No & 33.3 & 55.6 & 11.1 \\
Number & 48.3 & 37.9 & 13.8 \\
Other  & 17.2 & 48.3 & 34.5 \\
\bottomrule
\end{tabular}
\caption{Distribution (\%) of each VQA question type across the three processing modes.}
\label{tab:mode-by-category}
\end{table}

Table~\ref{tab:mode-by-category} shows a clear association between VQAv2 question types and the three modes. Yes/no questions are concentrated in Mode B (55.6\%), reflecting decisions that combine a localized target with broader contextual information. Number questions are split between Mode A (48.3\%) and Mode B (37.9\%): they fall into Mode A when counting can be resolved from localized instances, and into Mode B when it requires additional scene context beyond focal regions. Open-ended ``other'' questions are most frequently Mode B (48.3\%) and exhibit the highest proportion of Mode C (34.5\%), where answers are often driven by scene-level cues or image-conditioned knowledge rather than localized visual evidence. Overall, the mode structure aligns with the underlying reasoning demands induced by different question types.

\subsection{Stability Analyses for Three Modes}

\begin{table}[t]
\centering
\small
\setlength{\tabcolsep}{5pt}
\begin{tabular}{lccc}
\toprule
$n$ & A (\%) & B (\%) & C (\%) \\
\midrule
50 & 33.0 $\pm$ 4.2 & 47.0 $\pm$ 4.6 & 20.1 $\pm$ 3.7 \\
60 & 32.8 $\pm$ 3.2 & 47.1 $\pm$ 3.6 & 20.1 $\pm$ 2.8 \\
70 & 33.0 $\pm$ 2.4 & 47.0 $\pm$ 2.6 & 20.0 $\pm$ 2.0 \\
\midrule
\textit{\footnotesize 95\% CI} & {\footnotesize $[$23.9, 43.5$]$} & {\footnotesize $[$36.8, 57.6$]$} & {\footnotesize $[$12.9, 29.7$]$} \\
\bottomrule
\end{tabular}
\caption{\FV{Stability of the three-mode partition across sample sizes. Values are mean $\pm$ standard deviation; brackets denote the confidence intervals.}
}
\label{tab:mode-stability}
\end{table}

\FV{
As shown in Figure~\ref{fig:processing-modes-scatter} (b), the three-mode structure is already pronounced in the data. To quantify stability, we re-sample the valid samples without replacement at sizes at $n=50$, $60$, and $70$, repeating each 1000 times, and report Wilson 95\% confidence intervals on the full set in Table~\ref{tab:mode-stability}. The confidence intervals consistently exclude zero for all three modes, supporting the stability of the three-mode structure. Across sample sizes, repeated sub-sampling also yields highly consistent mode proportions, with the variance decreasing monotonically as $n$ increases. These results indicate that the three-mode partition is a data-driven characterization of this underlying distribution, rather than one induced by a particular sample size.
}

\subsection{GT versus Top-$k$ Divergence}
\label{sec:gt-topk}

\begin{figure}[t]
\centering
\includegraphics[width=0.96\columnwidth]{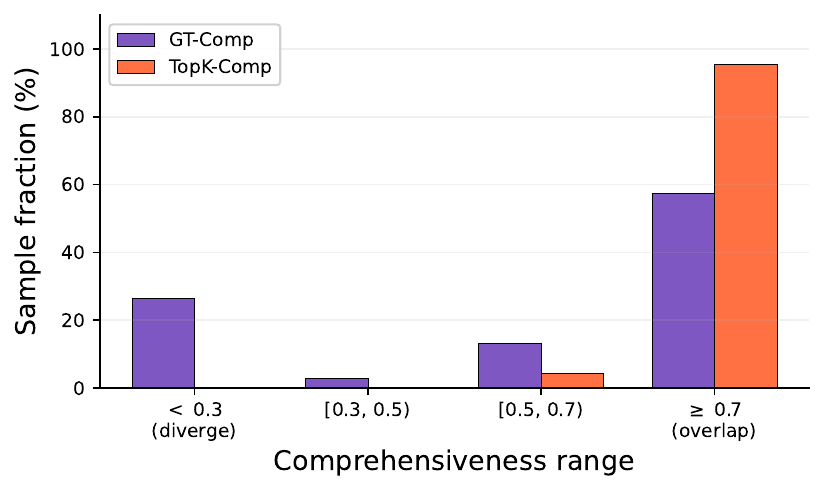}
\caption{Sample distribution of GT-Comp and Top-$k$ Comp in modes A and B.}
\label{fig:gt-topk-bimodal}
\end{figure}

Table~\ref{tab:strategy-aggregate} shows a substantial gap in comprehensiveness between the GT region and the overall top-$k$ region ($0.505$ vs.\ $0.813$), suggesting a mismatch between human-annotated evidence and the regions causally used by the model. We further analyze samples from Modes A and B, where overall top-$k$ comprehensiveness is consistently high. As shown in Figure~\ref{fig:gt-topk-bimodal}, GT comprehensiveness exhibits a bimodal distribution rather than concentrating at high values. The bimodal GT-Comp distribution separates samples where human-annotated regions align with the model’s causal evidence from those where they do not.

\begin{figure}[t]
\centering
\includegraphics[width=0.94\columnwidth]{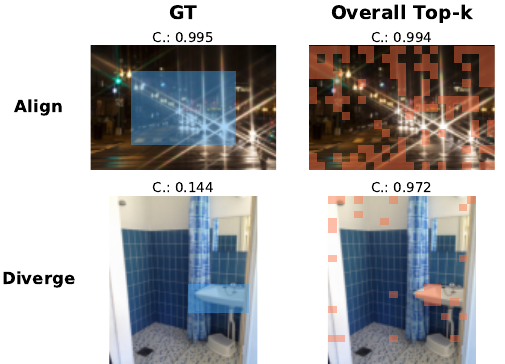}
\caption{Representative samples for the two GT versus Top-$k$ regimes. Top row: GT and Top-$k$ overlap on the salient answer object, and either ablation collapses the prediction. Bottom row: GT covers the named object but ablating it leaves the prediction nearly unchanged, while ablating the Top-$k$ region collapses it.}
\label{fig:gt-topk-examples}
\end{figure}

\begin{figure*}[t]
\centering
\begin{minipage}[c]{0.235\textwidth}
\centering
\includegraphics[height=3.5cm,keepaspectratio]{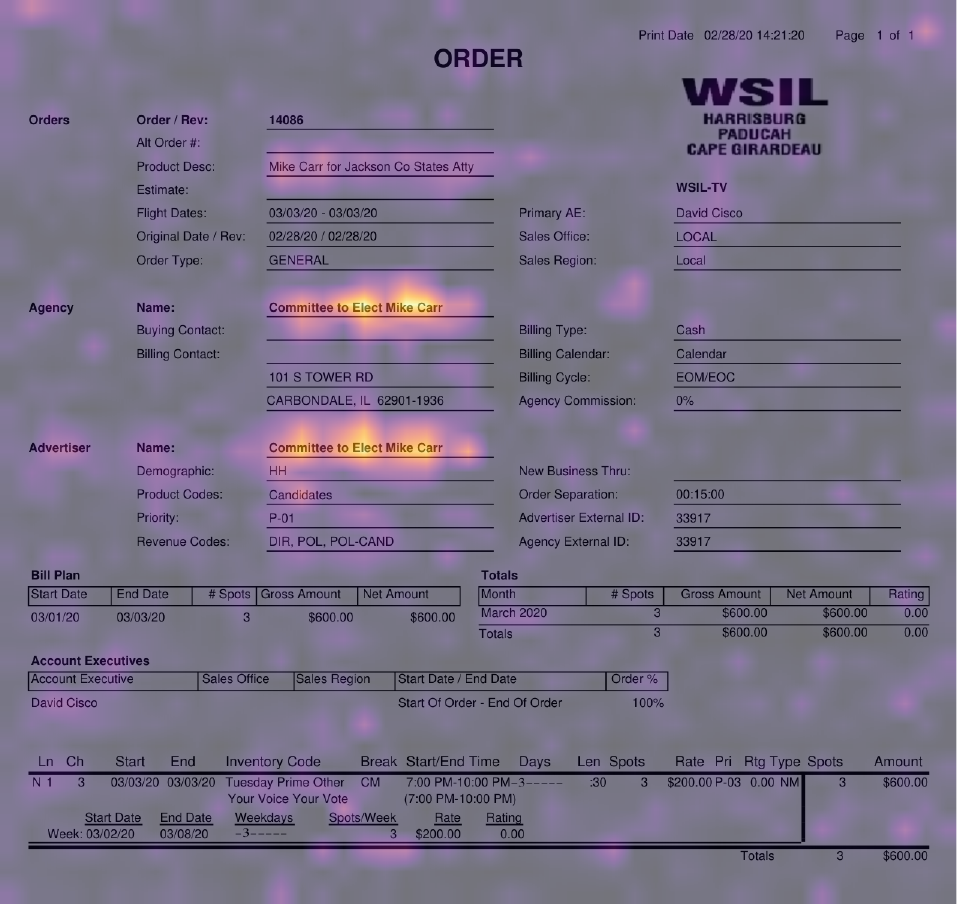}\\
{\footnotesize (a) Attention heatmap}
\end{minipage}\hspace{0.04\textwidth}%
\begin{minipage}[c]{0.56\textwidth}
\centering
\includegraphics[height=3.5cm,keepaspectratio]{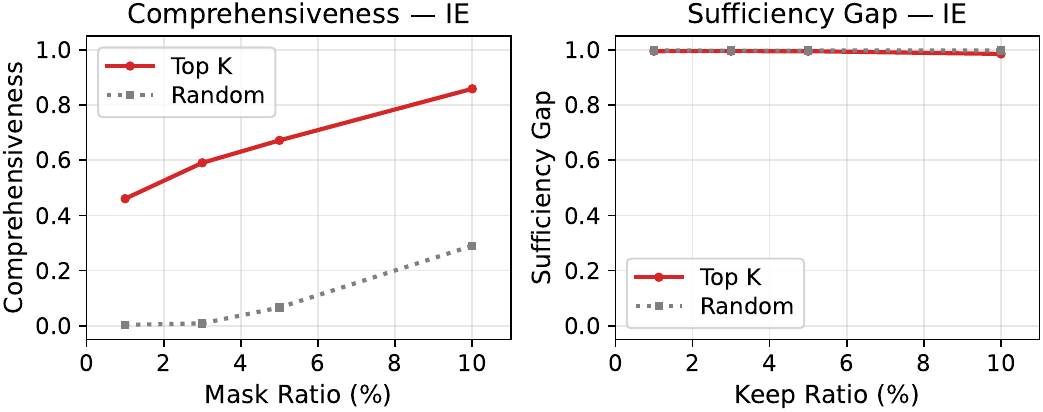}\\
{\footnotesize (b) Comp and Sgap vs.\ perturbation ratio}
\end{minipage}
\caption{(a) Attention concentrates on the answer string for the IE question \textit{Who is the advertiser?}. (b) Top-$k$ ablation (red) yields high Comp while a random baseline (grey) stays near zero; the Sgap remains near one for both, so the focal region is necessary but not sufficient for IE tasks on Qwen-VL-2.5.}
\label{fig:ie-overview}
\vspace{-0.3em}
\end{figure*}

In the aligned subset, mean GT-Comp reaches $0.963$, close to the mean Top-$k$ Comp of $0.992$, indicating that both regions capture the same causal signal, and \FV{the model answer accuracy is high at 97.4\%}. As shown in Figure~\ref{fig:gt-topk-examples} (top row), for the question \textit{Are there red lights?}, both the GT mask and the top-$k$ tokens concentrate on the lights, and ablating either region sharply reduces the answer probability. In contrast, the misaligned subset exhibits a mean GT-Comp of $-0.150$ while Top-$k$ Comp remains high at $0.945$, indicating that the human-annotated region contributes little to the prediction even when strong causal evidence is captured by the model’s top-$k$ regions, and \FV{the model answer accuracy is also high at 94.4\%}. Figure~\ref{fig:gt-topk-examples} (bottom row) shows a representative example for the question \textit{What color is the sink on the right?}: ablating the GT sink region leaves the prediction largely unchanged, whereas ablating the top-$k$ tokens substantially reduces the answer probability. This suggests that the model resolves the color attribute from surrounding contextual regions carrying correlated chromatic cues rather than from the object region itself.

These results highlight a key aspect of the model’s visual reasoning: while it exhibits partial alignment with human visual interpretation, it also relies on broader contextual dependencies beyond semantically corresponding regions. The model’s own attention more faithfully reflects its internal visual reliance than human-annotated GT regions. Therefore, treating GT masks from human understanding as a faithfulness ground truth for VLMs is insufficient.

\subsection{From VQA to Document \FV{Tasks}}
\label{sec:ie}

\begin{table}[t]
\centering
\small
\setlength{\tabcolsep}{7pt}
\begin{tabular}{lccc}
\toprule
\multirow{2}{*}{Strategy} & \multicolumn{3}{c}{Mask Ratio} \\
\cmidrule(lr){2-4}
 & 5\% & 10\% & 30\% \\
\midrule
Top-$k$ & \textbf{0.37\,/\,1.00} & \textbf{0.49\,/\,1.00} & \textbf{0.77\,/\,0.91} \\
Random  & 0.01\,/\,1.00 & 0.02\,/\,1.00 & 0.27\,/\,0.99 \\
\bottomrule
\end{tabular}
\caption{ChartQA causal perturbation on Qwen2.5-VL-7B. Each cell reports 
Comp\,/\,Sgap}
\label{tab:chartqa_perturb_qwen}
\end{table}

We further conduct the perturbation framework on document information extraction (IE) task, a practical application, where layout, repetition, and field structure replace natural-scene visual cues. Following the VQA formulation, each extraction target is cast as a question over the document, as illustrated in Figure~\ref{fig:ie-overview} (a). For example, for the query \textit{Who is the advertiser?} in a campaign-finance document, the attention heatmap concentrates on the answer string “Committee to Elect Mike Carr”, with minimal activation elsewhere on the page. This setting enables top-$k$-based perturbation analysis without relying on explicit GT region annotations.

As shown in Figure~\ref{fig:ie-overview} (b), ablating only the top $1\%$ of visual tokens already reduces the answer probability by $43.8\%$, and comprehensiveness increases to $0.826$ at the top $10\%$, indicating that a small focal region drives most of the prediction. In contrast, retaining only the top-$k$ tokens recovers at most $\sim 5\%$ of the prediction probability (Sgap $\sim0.947$), suggesting strong dependence on broader document context. Therefore, Qwen-VL-2.5 treats IE samples as an extreme instance of Mode B, where localized evidence is necessary but insufficient without layout-level context. 

\FV{For ChartQA samples in Table~\ref{tab:chartqa_perturb_qwen}, as the perturbation ratio increases, the top-$k$ comprehensiveness rises steadily and stays well above random, while the sufficiency gap remains close to 1, so the ChartQA task also falls into Mode B.}

These results extend the three-mode framework beyond VQA to document understanding, while suggesting practical implications for attention-aware token selection in VLM systems, where preserving broader layout context may remain essential.

\subsection{Cross-Model Generalizability}\label{sec:cross-model}
We further replicate the perturbation pipeline on InternVL2.5-8B, a VLM with a different LM backbone and a vision tower employing dynamic-tile preprocessing. 
On VQA, the overall top-$k$ visual tokens again attain the highest mean comprehensiveness ($0.733$) and the lowest mean sufficiency gap ($0.468$) among the four strategies, so attention-ranked tokens remain the most 
faithful selection%
; applying two-stage Otsu thresholding to InternVL's own distribution recovers the same three-mode partition, although the per-sample mode assignment overlaps only partially with Qwen's. On IE, attention is again concentrated on the answer string and comprehensiveness grows monotonically with the mask ratio as on Qwen, but the sufficiency gap decreases steadily and reaches $0.189$ at the top $10\%$, placing InternVL IE closer to mode A 
. \FV{On ChartQA, the trend is the same as the IE results and displays the mode A.}
These results suggest that heterogeneous faithfulness is a model-agnostic phenomenon, while which mode a particular task falls into depends on the underlying architecture, since different visual encoders and LM backbones may concentrate attention over different regions of the same image. Detailed 
results and discussion is provided in Appendix~\ref{app:internvl}.

\subsection{\FV{Scaling Across Model Sizes}}
\begin{table}[t]
\centering
\small
\setlength{\tabcolsep}{4pt}
\begin{tabular}{cccccc}
\toprule
 & \multicolumn{2}{c}{Comp} & \multicolumn{3}{c}{Mode (\%)} \\
\cmidrule(lr){2-3} \cmidrule(lr){4-6}
Model & Top-$k$ & GT & A & B & C \\
\midrule
Qwen2.5-VL-3B  & 0.739 & 0.432 & 31.8 & 37.6 & 30.6 \\
Qwen2.5-VL-7B  & 0.813 & 0.505 & 32.9 & 47.1 & 20.0 \\
\midrule
InternVL2.5-4B & 0.792 & 0.636 & 18.8 & 54.1 & 27.1 \\
InternVL2.5-8B & 0.733 & 0.609 & 29.4 & 45.9 & 24.7 \\
\bottomrule
\end{tabular}
\caption{\FV{Attention faithfulness across model scales.}}
\label{tab:scaling}
\end{table}

\FV{To assess the consistency of our findings across model scales, we replicate the VQA experiments on Qwen2.5-VL-3B and InternVL2.5-4B under the same protocol and compare them with their larger counterparts. As shown in Table~\ref{tab:scaling}, all three attention modes are observed across model sizes, and the attention-ranked top-$k$ regions consistently achieve higher comprehensiveness than the human-annotated GT regions. These results suggest that the observed faithfulness patterns are consistent across model scales.}
\vspace{-0.2em}
\section{Conclusion}
\vspace{-0.2em}
We systematically evaluate visual attention faithfulness in VLMs through token-level causal perturbation. Our results show that visual attention faithfulness is not binary, but instead manifests in three 
processing modes with different patterns of visual dependence, and different VLM architectures may process the same task under different modes. Furthermore, we find that human-annotated regions only align with the regions causally used by the model on some tasks, while the model relies more on hidden semantically corresponding regions for others. Experiments on document information extraction demonstrate that these patterns extend beyond general VQA to structured practical visual tasks. \XR{Together, these findings provide a unified perspective on how the visual attention faithfulness is manifested in VLMs.}
\section*{Limitations}
This work focuses on current VLMs with dynamic-resolution visual strategies, and may not directly extend to alternative 
schemes, such as models with fixed-resolution image inputs (e.g., LLaVA-style models). 
Second, we do not explicitly study reasoning-oriented VLMs 
that perform multi-step deliberation before answering, which may exhibit different visual attention behaviors, and we leave this direction for future work. 

\bibliography{main}

\appendix

\section{Details of Experiment Setup}
\label{app:setup}

\paragraph{Ground-Truth Region Annotation.}
For each of the 90 VQA samples, we define a ground-truth (GT) region as the set of visual tokens corresponding to the image area that a human would consider directly relevant to answering the question. Annotations are derived through a semi-automatic pipeline: we begin with COCO instance segmentation masks \cite{lin2014coco} overlaid on VQAv2 images, then manually select and refine bounding rectangles to cover exactly the task-relevant objects or areas (e.g., the road surface for ``How many lanes?'' or the traffic light for ``Is there a red light?''). Each annotation is expressed as a set of grid coordinates over the visual token grid, enabling direct correspondence between annotated regions and token positions. \FV{For the annotation process, first, a human expert, referencing the VQAv2 questions, images, and answers, and using COCO instance segmentation masks as a reference, selects the visual regions that can ultimately contribute to the answer. Then, a second human expert reviews the annotated image data in conjunction with the VQAv2 questions and answers, adjusting any that were inaccurate.}

\paragraph{Visual Token Grid.}
Qwen2.5-VL encodes each image into a spatial grid of visual tokens through patch embedding followed by $2\times2$ spatial merging, producing grid dimensions that vary with image resolution. Each language-visible token corresponds to a localized image region, and the flattened token ordering preserves the underlying 2D spatial arrangement through positional encoding. InternVL2.5 instead employs a dynamic tile-based preprocessor: the input image is partitioned into a variable number of $448{\times}448$ tiles together with a downscaled thumbnail tile. Each tile is independently encoded by the vision tower and transformed into a fixed-length token block via spatial downsampling, after which the per-tile blocks are concatenated and separated by tile-boundary tokens. The thumbnail tile provides an additional global-context representation alongside the locally constrained tile tokens. In both models, the language backbone receives a flat sequence of visual tokens whose ordering encodes the underlying spatial structure.

\section{Discussion}
\paragraph{Perturbation design.}
We perturb visual tokens at the entry of the first decoder layer of the language-model backbone, rather than masking image pixels or intervening inside the vision encoder. Since the object of evaluation is the language model's attention over visual tokens, the intervention is placed at the position where the language model reads those tokens, so that the measured probability change reflects the language model's reliance on a specific token rather than the vision encoder's response to a corrupted input. Pixel- or encoder-side ablations characterize the input dependencies of the vision encoder and offer a complementary view. 
Our setting is also align with recent token-level processing of visual tokens in VLMs \cite{chen2024fastv,shang2024prumerge,zhang2025sparsevlm}.

For the token-level perturbation, we replace the targeted hidden state with the zero vector instead of removing the token from the sequence: visual tokens occupy positions on a 2D spatial grid, and deleting a token would shift the positions of the remaining tokens and thereby distort the encoded layout, conflating the loss of a token's content with the loss of spatial structure. Zero-ablation preserves sequence length and positional structure while suppressing the content at the targeted positions, and is one of the standard activation-level interventions discussed alongside mean and Gaussian-noise ablation in activation-patching \cite{heimersheim2024patching}.

\begin{table}[t]
\centering
\footnotesize
\setlength{\tabcolsep}{3pt}
\resizebox{\columnwidth}{!}{%
\begin{tabular}{@{}lcccc@{}}
\toprule
\textbf{Sample} & \textbf{10\%} & \textbf{30\%} & \textbf{50\%} & \textbf{Mode} \\
\midrule
\textit{Is there a toy?}       & 1.00/1.00 & 0.99/1.00 & 1.00/0.99 & B \\
\textit{Is there rope around?} & 1.00/1.00 & 1.00/0.97 & 1.00/0.82 & B \\
\textit{Are these beignets?}   & 1.00/1.00 & 1.00/0.50 & 1.00/0.38 & B \\
\textit{How many snowboards?}  & 0.03/1.00 & 0.10/0.38 & 0.11/0.00 & C \\
\textit{What room is this?}    & 0.43/0.51 & 0.46/0.21 & 0.51/0.05 & C \\
\bottomrule
\end{tabular}%
}
\caption{Attention-ranked top-$k$ ablation on the five excluded VQA samples. Each cell reports $\text{Comp}/\text{Sgap}$.}
\label{tab:excluded-ratio}
\end{table}

\paragraph{Sample selection.}
We evaluate the framework on two task families that exercise different aspects of visual understanding. VQAv2 \cite{goyal2017vqav2} spans three question categories (yes/no, number, open-ended) associated with diverse visual-reliance patterns within a single benchmark. VRDU \cite{wang2023vrdu} complements VQAv2 with a structurally different visual regime, where text-rich layouts, field-level evidence, and naturally occurring redundancy test whether the framework transfers beyond natural images. Despite the relatively small evaluation scale, the three-mode structure exhibits clear separation in the $(\mathrm{Comp}, \mathrm{Sgap})$ space (Table~\ref{tab:mode-summary}), is reproduced on InternVL2.5-8B, and remains consistent for the five excluded VQA samples under a separate ratio-based protocol (Appendix~\ref{app:excluded-samples}). These observations suggest that the identified processing modes reflect stable behavioral patterns rather than dataset-specific artifacts.

\paragraph{Potential broader impact.}
The differences between Qwen2.5-VL (Mode B) and InternVL2.5 (closer to Mode A) on the same IE task suggest that different VLM backbones may adopt different visual processing preferences for the same task. This implies that visual attention faithfulness is not solely task-dependent but is partly model-dependent, which may affect the transferability of attention-aware token selection, compression, and interpretability methods across architectures. Our findings also suggest that human-annotated regions should not be treated as universal proxies for model visual reliance in real-world VLM applications.

\paragraph{\FV{Model selection rationale.}}
\FV{For model backbones, the two kinds of models, Qwen2.5VL and InternVL2.5, are chosen to represent two widely adopted paradigms for current dynamic-resolution VLMs. Qwen2.5-VL adopts native-resolution encoding, where the entire image is processed without explicit tiling, producing a variable number of visual tokens that scales with image resolution while preserving the original aspect ratio. InternVL2.5 adopts dynamic tiling, decomposing an image into a variable number of fixed-resolution tiles that are encoded separately. These together cover the mainstream design space of dynamic-resolution VLMs.}

\paragraph{\FV{Comparison with attention visualization.}}
\FV{Existing attention visualization methods primarily characterize where a model attends by presenting attention maps, whereas our work evaluates whether the attended regions faithfully explain the model's prediction. Our goal is therefore not to find a visualization method, but to explore the explanatory faithfulness of existing visual attention. Our causal evaluation further reveals cases where model-attended regions differ from human-annotated evidence in their contribution to the prediction, and enables the identification of distinct faithfulness modes that cannot be determined from attention maps alone.}

\paragraph{\FV{Relation to gradient-based saliency.}}
\FV{Gradient-based saliency and attention provide visual importance estimates from different signals: saliency is derived from gradients, whereas attention is obtained directly from the transformer's forward attention weights. Comparing their relative attribution quality addresses the question of attribution methods, whereas our focus is specifically on the faithfulness of visual attention itself in VLMs. We therefore directly evaluate attention through causal interventions, revealing heterogeneous faithfulness across samples and models. Extending the analysis to gradient-based attribution methods is complementary to the present study.}

\paragraph{\FV{Comparison with other work.}}
\FV{\citet{faithfulness-compare} also studies faithfulness in VLMs, but focuses on whether the explicit reasoning is grounded in the input image, relying on black-box methods such as self-reflection. In contrast, we investigate a mechanism-level question: whether the model's internal attention over visual tokens is faithful. We evaluate visual attention faithfulness through token-level ablation and quantify both comprehensiveness and sufficiency. Thus, while both works study visual faithfulness, we address a different object of faithfulness at a different level: we study internal attention faithfulness rather than the faithfulness of explicit reasoning.}

\begin{table}[t]
\centering
\small
\setlength{\tabcolsep}{4pt}
\begin{tabular}{lcccc}
\toprule
                  & \multicolumn{2}{c}{\textbf{Comp}} & \multicolumn{2}{c}{\textbf{Sgap}} \\
\cmidrule(lr){2-3}\cmidrule(lr){4-5}
\textbf{Strategy} & Qwen           & InternVL       & Qwen           & InternVL       \\
\midrule
GT Object         & 0.505          & 0.609          & 0.846          & 0.555          \\
Non-GT Top-$k$    & 0.659          & 0.474          & 0.653          & 0.659          \\
\textbf{Overall Top-$k$} & \textbf{0.813} & \textbf{0.733} & \textbf{0.469} & \textbf{0.468} \\
Random            & 0.502          & 0.487          & 0.599          & 0.549          \\
\bottomrule
\end{tabular}
\caption{Mean normalized comprehensiveness (Comp) and sufficiency gap (Sgap) of the four perturbation strategies on valid VQA samples.
Overall top-$k$ remains the strategy with the highest Comp and the lowest Sgap on both models.
}
\label{tab:internvl-strategy}
\end{table}

\begin{figure}[t]
\centering
\includegraphics[width=\columnwidth]{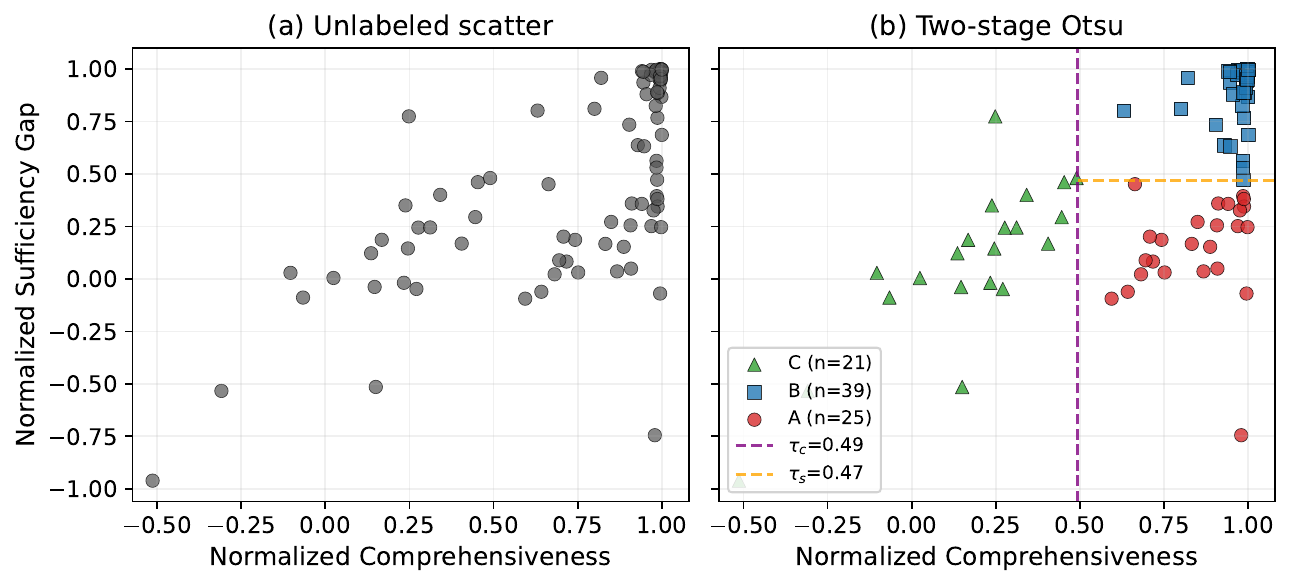}
\caption{Processing-mode scatter on InternVL2.5-8B.}
\label{fig:internvl-scatter}
\end{figure}

\begin{figure*}[t]
\centering
\begin{minipage}[c]{0.235\textwidth}
\centering
\includegraphics[height=3.5cm,keepaspectratio]{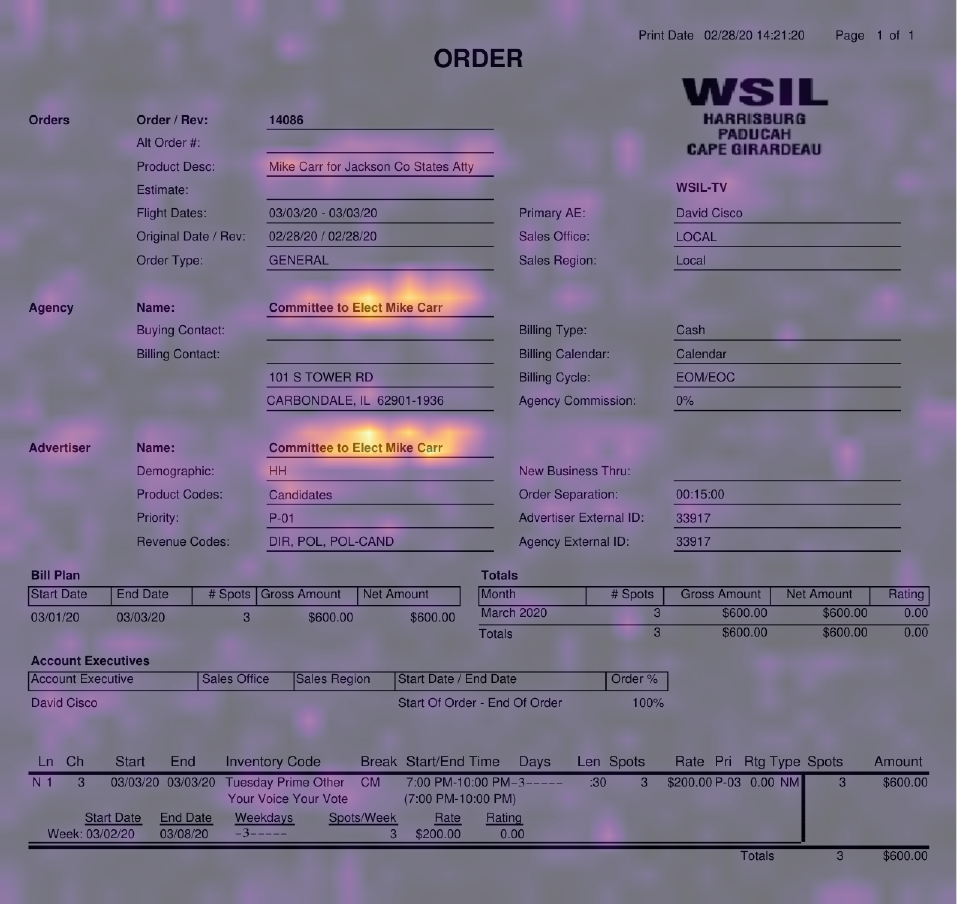}\\
{\footnotesize (a) Attention heatmap}
\end{minipage}\hspace{0.04\textwidth}%
\begin{minipage}[c]{0.56\textwidth}
\centering
\includegraphics[height=3.5cm,keepaspectratio]{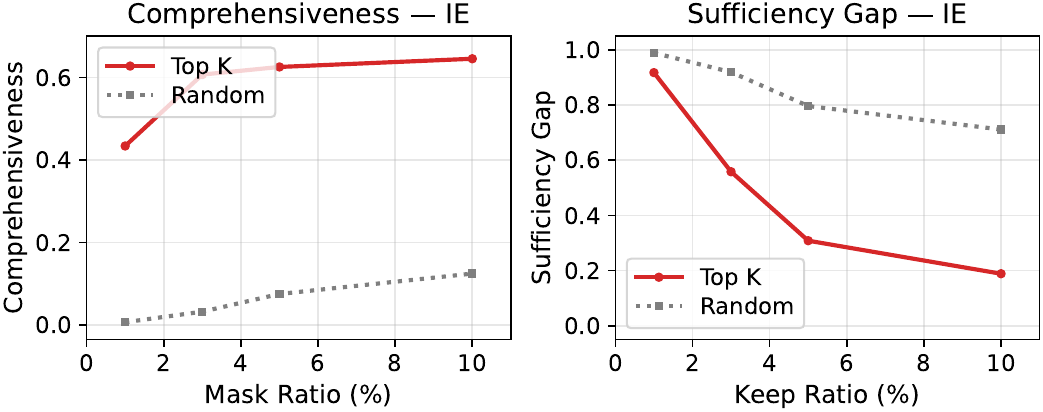}\\
{\footnotesize (b) Comp and Sgap vs.\ perturbation ratio}
\end{minipage}
\caption{(a) Aggregated attention on the advertiser sample concentrates on the answer string. (b) Comp rises and saturates while Sgap  decreases steadily and reaches $0.189$ at $k{=}10\%$, so the focal region is necessary and sufficient for IE tasks on InternVL2.5-8B.}
\label{fig:internvl-ie}
\end{figure*}
\section{Excluded VQA Samples}
\label{app:excluded-samples}

Five of the 90 collected VQA samples are excluded from the main analysis because their human-annotated GT regions cover almost the entire image, leaving little area for the GT-aligned strategies to be meaningfully different from full-image ablation. 
To further validate if our found faithfulness framework still works for these samples, we re-evaluate them on Qwen2.5-VL-7B using the ratio-based perturbation procedure: for $k\in\{10\%,30\%,50\%\}$ of all visual tokens, we mask the top-$k$ ranked by attention and measure Comp and Sgap directly. From Table~\ref{tab:excluded-ratio}, three samples fall into Mode~B (Faithful-Distributed): their Comp stays close to $1$ across all three perturbation ratios while Sgap remains high, so masking the attention-ranked top-$k$ collapses the prediction yet keeping only the top-$k$ does not restore it. The remaining two samples fall into Mode~C (Non-Focal): their Comp at the $50\%$ ratio is still low, indicating that removing the attention-ranked top-$k$ has a limited effect on the output.

For these samples, the B/C distribution aligns with the expectations of our faithfulness framework, as absence detection, counting ``0'', and holistic scene recognition draw their evidence from the whole image rather than from a key spatial subset. For faithfulness evaluation, the ratio-based protocol thus recovers the three-mode characterization on these samples even when the GT-aligned protocol does not apply. A systematic study of absence-detection and holistic-recognition questions in VQA is left for future work.

\section{InternVL2.5 Cross-Model Replication}
\label{app:internvl}

\begin{table}[t]
\centering
\small
\setlength{\tabcolsep}{7pt}
\begin{tabular}{lccc}
\toprule
\multirow{2}{*}{Strategy} & \multicolumn{3}{c}{Mask Ratio} \\
\cmidrule(lr){2-4}
 & 5\% & 10\% & 30\% \\
\midrule
Top-$k$ & \textbf{0.65\,/\,0.42} & \textbf{0.71\,/\,0.26} & \textbf{0.79\,/\,0.08} \\
Random  & 0.01\,/\,0.82 & 0.01\,/\,0.74 & 0.06\,/\,0.39 \\
\bottomrule
\end{tabular}
\caption{ChartQA causal perturbation on InternVL2.5-8B. Each cell reports Comp\,/\,Sgap. Retaining only the top-$30\%$ attention tokens already suffices (NSgap drops to $0.08$), placing chart understanding closer to the Faithful-Sufficient regime (mode A).}
\label{tab:chartqa_perturb_internvl}
\end{table}

We further validate our findings on InternVL2.5-8B, which combines a 32-layer InternLM2 language backbone with an InternViT-6B vision encoder and a tile-based visual preprocessor. Compared with Qwen2.5-VL's single-grid encoding, InternVL adopts a finer-grained tile-based strategy and typically produces a larger number of visual tokens for the same input image. Table~\ref{tab:internvl-strategy} compares the two models across the four perturbation strategies of \S\ref{sec:setup}; Figure~\ref{fig:internvl-scatter} shows the resulting processing-mode scatter; Figure~\ref{fig:internvl-ie} shows the IE replication on the same samples used in Figure~\ref{fig:ie-overview}.

On VQA, the overall top-$k$ visual tokens again attain the highest mean comprehensiveness ($0.733$) and the lowest mean sufficiency gap ($0.468$) among the four strategies, so attention-ranked tokens remain the most faithful selection on InternVL. The InternVL Comp value is below the Qwen counterpart ($0.733$ vs.\ $0.813$), suggesting that the same fraction of attention-top tokens carries less unique necessity signal under the dynamic-tile encoder, and that visual information is more redundantly distributed across visual tokens. Two-stage Otsu thresholding on InternVL's own distribution recovers the same three-mode partition (Figure~\ref{fig:internvl-scatter}); the resulting $\tau_c{=}0.49$ falls below Qwen's $\tau_c{=}0.59$, consistent with this redundancy reading. Per-sample mode overlap with the Qwen labels (\S\ref{sec:three-modes}) is $50.6\%$: the same VQA item may activate different attention regions across models, while the coexistence of the three modes is model-agnostic. On IE (Figure~\ref{fig:internvl-ie}), Comp continues to rise with the mask ratio as on Qwen and saturates near $0.650$ by $k{=}3\%$, while Sgap drops steadily to $0.189$ at $k{=}10\%$, and \FV{the trend is the same on the ChartQA task as shown in table~\ref{tab:chartqa_perturb_internvl}}, placing InternVL's IE \FV{and ChartQA tasks} in mode A rather than the mode B regime observed on Qwen.

In Qwen2.5-VL, the top-$k$ visual tokens tend to provide partial but insufficient evidence, as the representation is distributed across spatial regions and requires broader contextual aggregation for layout reconstruction. In contrast, InternVL decomposes the image into multiple localized tiles together with a global thumbnail tile, making local visual evidence more spatially concentrated while introducing an additional compact global-context representation. As a result, a small subset of selected tokens can often form a near-sufficient evidence set for answer recovery. This design choice may make IE attention in InternVL behave more like a near-sufficient set than a distributed signal, explaining why InternVL IE converges to mode A whereas Qwen IE remains in mode B. Despite these differences in visual encoding, the three-mode framework itself transfers across both architectures, indicating that it characterizes how a VLM allocates visual attention at the language-model side rather than a property tied to a specific vision encoder.

\section{Usage of AI Assistant}
We use AI assistants or tools such as Claude and Grammarly to correct grammar errors and polish the language.

\end{document}